%% file: partialTraceRegression.tex
\documentclass{article}

\usepackage{microtype}
\usepackage{graphicx}
\usepackage{subfigure}
\usepackage{booktabs} 

\usepackage{hyperref}

\usepackage[accepted]{icml2020}

\usepackage[utf8]{inputenc} 
\usepackage[T1]{fontenc}    
\usepackage{hyperref}       
\usepackage{url}            
\usepackage{booktabs}       
\usepackage{amsfonts}       
\usepackage{nicefrac}       
\usepackage{microtype}      

\usepackage{amsmath}
\usepackage{amssymb}
\usepackage{csquotes}
\usepackage{graphicx}
\usepackage{wrapfig}
\usepackage{bbm}
\usepackage{todonotes}

\newtheorem{de}{Definition}
\newtheorem{theo}{Theorem}

\newtheorem{lem}[theo]{Lemma}
\newtheorem{remark}{Remark}

\DeclareMathOperator*{\argmin}{arg\,min}

\usepackage{color}
\definecolor{MyDarkBlue}{rgb}{0.05,0.,0.8}

\newcommand{\tr}{\text{tr}}

\newcommand{\symmetric}{\mathbb{S}}
\newcommand{\matrices}{\mathbb{M}}

\icmltitlerunning{Partial Trace Regression and Low-Rank Kraus Decomposition}

\begin{document}
	
	\twocolumn[
	\icmltitle{Partial Trace Regression and Low-Rank Kraus Decomposition}
	
	
	
	\icmlsetsymbol{equal}{*}
	
	\begin{icmlauthorlist}
		\icmlauthor{Hachem Kadri}{amu}
		\icmlauthor{St\'ephane Ayache}{amu}
		\icmlauthor{Riikka Huusari}{aalto}
		\icmlauthor{Alain Rakotomamonjy}{ur,criteo}
		\icmlauthor{Liva Ralaivola}{criteo}
	\end{icmlauthorlist}
	
	\icmlaffiliation{amu}{Aix-Marseille University, CNRS, LIS, Marseille, France}
	\icmlaffiliation{aalto}{Helsinki Institute for Information Technology HIIT, Department of Computer Science, Aalto University, Espoo, Finland}
	\icmlaffiliation{ur}{Universit\'e Rouen Normandie, LITIS, Rouen, France}
	\icmlaffiliation{criteo}{Criteo AI Lab, Paris, France}
	
	\icmlcorrespondingauthor{Hachem Kadri}{hachem.kadri@univ-amu.fr}
	
	\icmlkeywords{Trace Regression, Partial Trace, Completely Positive Maps, Kraus Decomposition, Matrix Completion}
	
	\vskip 0.3in
	]
	
	
	
	\printAffiliationsAndNotice{}  

\begin{abstract}
	The trace regression model, a direct extension of the well-studied linear regression model, allows one
	to map matrices to real-valued outputs. We here introduce an even more general model, namely the
	{\em partial-trace regression model}, a family of linear mappings from matrix-valued
	inputs to {\em matrix-valued outputs}; this model subsumes the trace regression model and thus the linear regression model. Borrowing tools from quantum information theory, where partial trace
	operators have been extensively studied, we propose a framework for learning partial trace
	regression models from data by taking advantage of the so-called low-rank Kraus
	representation of {\em completely positive maps}. We show the relevance of our framework with
	synthetic and real-world experiments  conducted for  both i) matrix-to-matrix regression and ii) positive semidefinite matrix completion, 
	two tasks which can be formulated as partial trace regression problems.
	 
\end{abstract}

\section{Introduction}
\label{sec:intro}
\input{intro.tex}


\section{Partial Trace Regression}
\label{sec:ptr}
\input{ptr.tex}

\section{Experiments}
\label{sec:expe}
\input{expe.tex}

\section{Conclusion}
\label{sec:conc}
\input{conclusion.tex}

\section*{Acknowledgements}
This work has been funded by the French National Research Agency (ANR) project QuantML (grant number ANR-19-CE23-0011).  Part of this work  was performed using computing resources of CRIANN (Normandy, France). 
The work by RH has in part been funded by Academy of Finland grants 334790 (MAGITICS) and 310107 (MACOME).

\bibliography{ref}

\bibliographystyle{icml2020}

\newpage
\section*{Supplementary Material}
\label{sec:supp}
\input{appendix.tex}

\end{document}

%% file: intro.tex

\textbf{Trace regression model. \ }
The {\em trace regression model} or, in short, {\em trace regression}, is a 
generalization of the well-known linear regression model to the case where input data are matrices instead of vectors~\cite{rohde2011estimation, koltchinskii2011nuclear, slawski2015regularization}, with the output still being real-valued. This model assumes, for the pair of covariates $(X,y)$,
 the following relation between the matrix-valued random variable $X$ and the real-valued random variable $y$:
\begin{equation}
\label{eq:trModel}
y = \tr\left(B_*^{\top} X\right) + \epsilon,
\end{equation}
where $\tr(\cdot)$ denotes the trace,  $B_{*}$ 
is some unknown matrix of regression coefficients, and $\epsilon$ is random noise.
This model has been used beyond mere matrix regression
for problems such as phase retrieval~\cite{candes2013phaselift}, quantum state tomography~\cite{gross2010quantum},
 and matrix completion~\cite{klopp2011rank}.
 
 \begin{figure}[t]
 	\centering
 	\includegraphics[width=0.4\textwidth]{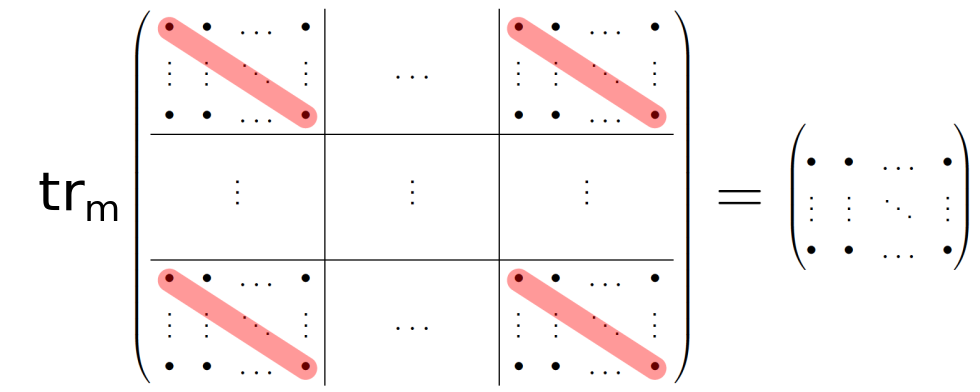}
 	\caption{\label{fig:pt}Illustration of the partial trace operation. The partial trace operation applied to $m\times m$-blocks of a $qm\times qm$ matrix gives a $q\times q$ matrix as an output. }
 \end{figure}

Given a sample $S=\{(X_i,y_i)\}_{i=1}^\ell$, where $X_i$ is a ${p_1\times p_2}$ matrix and  $y_i\in \mathbb{R}$, and each $(X_i,y_i)$ is
assumed to be distributed as $(X,y)$, the training task associated
with statistical model~\eqref{eq:trModel} is to 
find a matrix $\widehat{B}$ that is a proxy to $B_*$.
To this end, \citet{koltchinskii2011nuclear, fan2019generalized} proposed to compute an estimation $\widehat{B}$ of $B_*$  as 
the solution of the regularized least squares problem
\begin{equation}
\label{eq:trOptim}
\widehat{B} = \argmin_{B}\sum_{i=1}^\ell \left(y_i -  \tr\left(B^{\top}X_i\right)\right)^2 + \lambda \|B\|_1,
\end{equation}
where $\|\cdot\|_1$ is the trace norm (or nuclear norm), which promotes a low-rank $\widehat{B}$, a key feature for the authors to establish bounds on the deviation
of $\widehat{B}$ from $B_*$. \citet{slawski2015regularization, koltchinskii2015optimal} have considered the particular case where $p\doteq p_1=p_2$ and $B_*$ is
assumed to be from $\symmetric_p^+$, the cone of positive semidefinite matrices of order $p$, and they showed that guarantees on the deviation 
of $\widehat{B}$ from $B_*$  continue to hold when $\widehat{B}$ is computed as
\begin{equation}
\label{eq:trPSD}
\widehat{B} = \argmin_{B\in\symmetric_p^{+}}\sum_{i=1}^\ell \left(y_i -  \tr\left(B X_i\right)\right)^2.
\end{equation}
Here, the norm regularization of~\eqref{eq:trOptim} is no longer present and it is replaced
by an explicit restriction for $\widehat{B}$ to be in $\symmetric_p^{+}$~(as $B_*$).
This setting is tied to the learning of {\em completely positive maps} developed hereafter.

\textbf{Partial trace regression model. \ }
Here, we propose the partial trace regression model, that generalizes the trace regression model to the case when both inputs and outputs are matrices,
and we go a step farther from works that are assuming either matrix-variate inputs \cite{zhou2014regularized, ding2014dimension, slawski2015regularization, luo2015support} or matrix/tensor-variate  outputs~\cite{kong2019l2rm, li2017parsimonious, rabusseau2016low}. Key to our work is
the so-called partial trace, explained in the following section and depicted in Figure~\ref{fig:pt}.

This novel regression model that maps matrix-to-matrix is of interest for several
application tasks. For instance, in Brain-Computer Interfaces, covariance matrices are frequently used as a feature for representing mental state of a subject~\cite{barachant2011multiclass,CongedoBA13} and those covariance matrices need
to be transformed \cite{ZaniniCJSB18} into other covariance matrices to be discriminative for some BCI tasks. Similarly in Computer Vision and especially 
in the subfield of 3D Shape retrieval, covariance matrices are of interest as descriptors \cite{GuoWX18,HaririTFBD17}, while there is a surging interest in deep learning methods for defining trainable layers with covariance matrices as input and output \cite{huang2017riemannian,BrooksSBSC19}.

\textbf{Contributions.\ } We make the following contributions. 
i)~We introduce 
the
{\em partial trace regression model}, a family of linear predictors from matrix-valued inputs to matrix-valued outputs; this model encompasses 
previously known regression models, including  the trace regression model and thus the linear regression model;
ii)~borrowing concepts from quantum information theory, where partial trace operators have been extensively studied, we propose a framework for learning a
partial trace regression model from data; we take advantage of  the low-rank Kraus representation of completely positive maps  to cast learning as an 
optimization problem which we are able to handle efficiently;
iii)~we  provide statistical guarantees for the model learned under our framework, thanks to a provable estimation of pseudodimension of the class of
functions that we can envision;
iv) finally, we show the relevance of the proposed framework for the tasks of matrix-to-matrix regression and positive semidefinite matrix completion, both of them are amenable to a partial trace regression formulation; our empirical results show that partial trace regression model yields good performance, demonstrating wide applicability and effectiveness.


%% file: ptr.tex


Here, we introduce the partial trace regression model to encode linear mappings from matrix-valued spaces to matrix-valued spaces. 
We specialize this setting to completely positive maps, and show the optimization problem to which learning with the partial trace regression model translates,
together with a generalization error bound for the associated learned model. In addition, we present how the problem of~(block) positive semidefinite matrix completion 
can be cast as a partial trace regression problem.

\begin{table}[t]
	\centering
	\begin{tabular}{cl}
		\toprule
		Symbol & \multicolumn{1}{c}{Meaning}\\
		\midrule
																			
		 $i$, $j$, $m$, $n$, $p$, $q$                            &           integers                   \\ 
		 $\alpha$, $\beta$, $\gamma$, $\ldots$                   &           real numbers               \\ 
		$\mathcal{X}$, $\mathcal{Y}$, $\mathcal{H}$, $\ldots$    &           vector spaces\footnotemark \\ 
		$x$, $y$, $k$, $\ldots$                                  &            vectors (or functions)     \\ 
		$X$, $Y$, $K$, $\ldots$                                  &            matrices (or operators)    \\ 
		$\mathbf{X}$, $\mathbf{Y}$, $\mathbf{K}$, $\ldots$       &            block  matrices            \\ 
		$\Phi$, $\Lambda$, $\Gamma$, $\ldots$                    &              linear maps on matrices    \\ 
			$\top$                                   & transpose   \\
		\bottomrule
	\end{tabular}
	\caption{Notational conventions used in the paper.}
	\label{tab:notation}
\end{table}


\subsection{Notations, Block Matrices and Partial Trace}  
\paragraph{Notations.} Our notational conventions are summarized in Table~\ref{tab:notation}. For $n,m \in \mathbb{N}$, $\mathbb{M}_{n \times m} = \mathbb{M}_{n \times m}(\mathbb{R})$ denotes the space of all $n \times m$ real matrices. If $n = m$, then we write $\mathbb{M}_{n}$ instead of $\mathbb{M}_{n \times n}$.
If $M$ is a matrix, $M_{ij}$ denotes its $(i,j)$-th entry.  
For $M\in\mathbb{M}_{n}$, $M \succeq 0$ will be used to indicate that $M$ is positive semidefinite (PSD); we may equivalently write $M\in\symmetric_n^+$.
Throughout, $\{(X_i,Y_i)\}_{i=1}^l$  denotes a training sample of $l$ examples, with each $(X_i,Y_i)$ assumed to be drawn IID from a fixed but unknown distribution
on $\mathcal{X} \times \mathcal{Y}$ where, from now on, $\mathcal{X}\doteq\mathbb{M}_{p}$ and $\mathcal{Y}\doteq\mathbb{M}_{q}$.

\paragraph{Block matrices.} We will make extensive use of the notion of block matrices, i.e., matrices that
can be partitioned into submatrices of the same dimension. If $M\in\matrices_{nm}$, the number of block partitions of $M$ directly
depends on the number of factors of $n$ and $m$; to uniquely identify the partition we are working with, we will always
consider a $n\times n$ block partition, where $n$ will be clear from context ---the number of rows and 
columns of the matrix at hand will thus be multiples of $n$. The set $\mathbb{M}_{n}(\mathbb{M}_{m})$ will then
denote the space of $n\times n$ block matrices $\mathbf{M}=[[\mathbf{M}_{ij} ]]$ whose $i, j$ entry is an element of  $\mathbb{M}_m $.\footnotemark

\setcounter{footnote}{1}
\footnotetext{We also use the standard notations such as $\mathbb{R}^n$ and $\mathbb{M}_n$.}
\addtocounter{footnote}{1}

\footnotetext{The space $\mathbb{M}_{n}(\mathbb{M}_{m})$ is isomorphic to $\mathbb{M}_{n} \otimes \mathbb{M}_{m}$. 
}

\paragraph{Partial trace operators.}
Partial trace, extensively studied and used in quantum computing~(see, e.g,. \citealt[Chapter~10]{rieffel2011quantum}), 
generalizes the trace operation to block matrices. The definition we work with is the following.

\begin{de}\label{de:partialTrace} (Partial trace, see, e.g.,~\citealp{bhatia2003partial}.)
 The \textit{partial trace} operator, denoted by $\mathrm{tr}_m(\cdot)$, is the linear map from $\mathbb{M}_{q}(\mathbb{M}_{m})$ to $\mathbb{M}_{q}$ defined by
	\begin{equation*}
\mathrm{tr}_m(\mathbf{M}) = \big(\mathrm{tr}(\mathbf{M}_{ij})\big), i,j=1,\ldots,q.
\end{equation*}
\end{de}
In other words, given a block matrix $\mathbf{M}$ of size $qm \times qm$, the partial trace is  obtained by computing the trace of each block of size $m\times m$ in the input matrix $\mathbf{M}$, as depicted in Figure~\ref{fig:pt}. We note that in the particular case $q=1$, the partial trace {\em is} the usual trace. 

\begin{remark}[Alternative partial trace]
	The other way of generalizing the trace operator to block matrices is the so-called block trace~\cite{filipiak2018properties}, 
	which sums the diagonal blocks of block matrices. We do not use it here. 
\end{remark}

\subsection{Model}

We now are set to define the {\em partial trace regression} model. 
\begin{de}\label{de:ptrModel} (Partial trace regression model)
	The partial trace regression model assumes for a matrix-valued
	covariate pair $(X,Y)$, with $X$ taking value in $\matrices_{p}$ 
	and $Y$ taking value in $\matrices_q$:
	\begin{equation}
		\label{eq:ptrModel}
		Y= \mathrm{tr}_m\left(A_* X B_*^{\top}\right) + \epsilon,
	\end{equation}
	where $A_*,B_*\in\mathbb{M}_{qm\times p}$ are the unknown parameters of the model and $\epsilon$ is some matrix-valued random noise.
\end{de}	

This assumes a stochastic linear relation between the input $X$ and the corresponding output $Y$ that,
given an IID training sample $\{(X_i,Y_i)\}_{i=1}^l$ drawn according to~\eqref{eq:ptrModel}, points
to the learning of a linear mapping $\widehat{\Phi}:\matrices_p\to\matrices_q$ of the form
\begin{equation}
	\label{eq:generalphi}
	\widehat{\Phi}(X)=\tr_m\left(\widehat{A}X\widehat{B}^{\top}\right),
\end{equation}
where $\widehat{A},\widehat{B}\in\matrices_{qm\times p}$ are the parameters to be estimated.

When $q=1$, we observe that $\mathrm{tr}_m(A_* X B_*^{\top}) = \mathrm{tr}(A_* X B_*^{\top})=\mathrm{tr}( B_*^{\top}A_* X)$, which
is exactly the trace regression model~\eqref{eq:trModel}, with a parametrization of the regression matrix as $B_*^{\top}A_*$.

We now turn our attention to the question as how to estimate the matrix parameters of the partial trace regression model
while, as was done in~\eqref{eq:trOptim} and~\eqref{eq:trPSD}, imposing some structure on the estimated parameters, so for the
learning to come with statistical guarantees. As we will see, our solution to this problem takes inspiration 
from the fields of quantum information and quantum computing, and amounts to the use of the Kraus representation of
 completely positive maps.

\subsection{Completely Positive Maps, Kraus Decomposition}
The space $\mathcal{L}(\mathbb{M}_{p},\mathbb{M}_{q})$ of linear
maps from $\mathbb{M}_{p}$ to $\mathbb{M}_{q}$ is a real vector space that has been
thoroughly studied in the fields of mathematics, physics, and more 
specifically quantum computation and information~\cite{bhatia2009positive, nielsen2000quantum, stormer2012positive, watrous2018theory}.

Operators from $\mathcal{L}(\mathbb{M}_{p},\mathbb{M}_{q})$ that have special properties
are the so-called completely positive maps, a family that builds upon the 
notion of positive operators.
\begin{de}(Positive maps, \citealp{bhatia2009positive}) A linear map
	$\Phi\in\mathcal{L}(\mathbb{M}_{p},\mathbb{M}_{q})$ is {\em positive} 
	if for all $M\in\symmetric_p^+$, $\Phi(M)\in\symmetric_q^+$.
\end{de}

To define completely positive maps, we are going to
deviate a bit from the block matrix structure advocated before and
 consider block matrices from $\matrices_m(\matrices_p)$
and $\matrices_m(\matrices_q)$.
\begin{de}(Completely positive maps, \citealp{bhatia2009positive})
	$\Phi\in\mathcal{L}(\mathbb{M}_{p},\mathbb{M}_{q})$ is {\em $m$-positive}
 if the associated map $\Phi_m\in\mathcal{L}(\matrices_m(\mathbb{M}_{p}),\matrices_m(\mathbb{M}_{q}))$ 
	which computes the $(i,j)$-th block of $\Phi_m({\bf M})$ as $\Phi({\bf M_{ij}})$ is positive. 
	
	$\Phi$ is {\em completely positive} if it is $m$-positive for any $m\geq 1.$
\end{de}

The following theorem lays out the connection between partial trace regression models and positive maps.
\begin{theo}\label{theo:stinespring}{(Stinespring representation,~\citealp{watrous2018theory})}
	Let $\Phi \in \mathcal{L}(\mathbb{M}_{p},\mathbb{M}_{q})$.
	$\Phi$ writes as $\Phi(X)=\mathrm{tr}_m\left(A X A^{\top}\right)$ for some $m\in\mathbb{N}$ and $A\in\mathbb{M}_{qm\times p}$ {\em if and only if} $\Phi$ is completely positive.
\end{theo}
This invites us to solve the partial trace learning problem by looking for a map $\widehat{\Phi}\in\mathcal{L}(\mathbb{M}_{p},\mathbb{M}_{q})$
that writes as:
\begin{equation}
	\label{eq:cpphi}
	\widehat{\Phi}(X)=\tr_m(\widehat{A}X\widehat{A}^{\top}),
\end{equation}
where now, in comparison to the more general model of~\eqref{eq:generalphi}, the operator $\widehat{\Phi}$ to be
estimated is a completely positive map that depends on a sole matrix parameter $\widehat{A}$. Restricting ourselves
to such maps might seem restrictive but i) this is nothing but the partial trace version of the PSD contrained trace regression
model of~\eqref{eq:trPSD},  which allows us to establish statistical guarantees, ii) the entailed optimization problem
can take advantage of the Kraus decomposition of completely positive maps (see below) and iii) empirical performance
is not impaired by this modelling choice.

Now that we have decided to focus on learning completely positive maps, 
we may introduce the last ingredient of our model, the Kraus representation.

\begin{theo}\label{theo:kraus}{(Kraus representation,~\citealp{bhatia2009positive})}
	Let $\Phi\in\mathcal{L}(\mathbb{M}_{p},\mathbb{M}_{q})$ be a completely positive linear map. Then there exist $A_j\in\mathbb{M}_{q\times p}$, $1\leq j \leq r$, with  $r\leq pq$ such that
	\begin{equation}\label{eq:kraus}
		\forall X\in \mathbb{M}_{p},\quad \Phi(X) = \sum_{j=1}^{r} A_j X A_j^\top.
	\end{equation}
\end{theo}

The matrices $A_j$ are called Kraus operators and $r$ the Kraus rank of $\Phi$.

With such a possible decomposition, learning a completely positive map  $\widehat{\Phi}$ can reduce to finding Kraus operators $A_j$,
for some fixed hyperparameter $r$, where small values of $r$ correspond to low-rank Kraus decomposition~\citep{aubrun2009almost,lancien2017approximating}, and favor
 computational efficiency and statistical guarantees. Given a sample $\{(X_i,Y_i)\}_{i=1}^l$, the training problem can now
be written as:
\begin{equation}
\label{eq:ptrOpt2}
\argmin_{A_j \in\mathbb{M}_{q\times p}} \sum_{i=1}^l  \ell\big(Y_i,\sum_{j=1}^r A_j X_i A_j^\top\big),
\end{equation}
where $\ell$ is a loss function.  The loss function we use in our experiments  is the square loss $\ell(Y,\widehat{Y}) = \|Y - \widehat{Y}\|_F^2$, where $\|\cdot\|_F$ is the Frobenius norm.
When $\ell$ is the square loss and $q=1$, problem~(\ref{eq:ptrOpt2}) reduces to the PSD contrained trace regression problem~(\ref{eq:trPSD}) , with a parametrization of the regression matrix as $\sum_{j=1}^r  A_j^\top A_j $.

\begin{remark}
	Kraus and Stinespring representations can fully characterize completely positive maps. It has been shown that for a Kraus representation of rank $r$, there exists a Stinespring representation with dimension $m$ equal to $r$~\citep[Theorem~2.22]{watrous2018theory}. 
	Note that the Kraus representation is rather computationally friendly compared to Stinespring representation. It has a simpler form and is easier to store and manipulate, as no need to create the matrix $A$ of size $qm\times p$.
	It also allows us to derive a generalization bound for partial trace regression, as we will see later.
\end{remark}

\smallskip

\textbf{Optimization. \ }Assuming that the loss function $\ell(\cdot,\cdot)$ is convex in its second argument, the resulting
objective function is non-convex with respect to $A_j$. If we further assume that the loss
is coercive and differentiable, then the learning problem admits a solution that is potentially a local minimizer.  In practice, several classical approaches
can be applied to solve this problem. We have for instance tested a block-coordinate
descent algorithm \cite{luo1992convergence} that optimizes one $A_j$ at a time. However, given current algorithmic tools, we have opted to solve the learning problem \eqref{eq:ptrOpt2} using autodifferentiation \cite{baydin2017automatic} and stochastic gradient descent, since the model can be easily implemented as a sum of product of matrices. This has the advantage of being efficient and allows one to leverage on
efficient computational hardware, like GPUs.

Note that at this point, although not backed by theory, we can consider
multiple layers of mappings by composing several mappings $\{\Phi_k\}$.
This way of composing  would extend the BiMap layer introduced by \citet{huang2017riemannian} which limits their models to
rank $1$ Kraus decomposition. Interestingly, they also proposed a ReLU-like
layer for PSD matrices that  can be applicable to our work as well. For two layers, this would boil down to  $\Phi_2 \circ \Phi_1(X) = \sum_{j_2=1}^{r_2} A^{(2)}_{j_2} \ \Gamma[\sum_{j_1=1}^{r_1} A^{(1)}_j X A_j^{(1)^\top}] \ A_{j_2}^{(2)^\top}$, where $\Gamma$ is a nonlinear activation that preserves positive semidefiniteness. We investigate also this direction in our experiments.

\smallskip

\input{bound.tex}

\subsection{Application to PSD Matrix Completion}

Our partial trace model is designed to address the problem of  matrix-to-matrix regression. We now show how it can also be applied to the problem of matrix completion.
We start by recalling how the matrix completion problem fits into standard trace regression model. Let $B^*\in\mathbb{M}_{m}$ be a matrix whose entries $B^*_{ij}$ are given only for some $(i,j)\in \Omega$. 
  Low-rank matrix completion can be addressed by solving the following optimization problem:
 \begin{equation}
 \label{eq:matrixcompletion}
 \argmin_{B} \|\mathcal{P}_\Omega(B) - \mathcal{P}_\Omega(B^*)\|^2, \text{  s.t. } \mathrm{rank}(B)= r,
 \end{equation}
where $\mathcal{P}_\Omega(B)_{ij} = B_{ij}$ if $(i,j) \in \Omega$, 0 otherwise. This problem can be cast as a trace regression problem by considering $y_{ij} = \mathcal{P}_\Omega(B^*)_{ij}$ and $X_{ij} = E_{ij}$, where $E_{ij}$, $1\leq i,j\leq m$, are the matrix units of  $\mathbb{M}_m$. 
Indeed, it is easy to see that in this case  problem~(\ref{eq:matrixcompletion}) is equivalent  to 
 \begin{equation}
\label{eq:matrixcompletion1}
\argmin_{B} \sum_{(i,j)\in \Omega} \big(y_{ij} - \mathrm{tr}(BX_{ij})\big)^2 \text{  s.t. } \mathrm{rank}(B)= r.
\end{equation}

Since the partial trace regression is a generalization of the trace regression model, one can ask what type of  matrix completion problems can be viewed as partial trace regression models. The answer to this question is given by the following theorem.
\begin{theo}\label{theo:blockcompletion}{(\citealp[Theorem~2.49]{hiai2014introduction})}\\[0.1cm]
	Let $\Phi : \mathbb{M}_{p} \to \mathbb{M}_{q}$ be a linear mapping. Then the following conditions are equivalent:\\[-0.7cm]
	\begin{enumerate}
		\item $\Phi$ is completely positive.
		\item The block matrix $\mathbf{M} \in \mathbb{M}_p(\mathbb{M}_q)$ defined by
			\begin{equation}
		\mathbf{M}_{ij} = \Phi(E_{ij}), \    1\leq i, j \leq p,
		\end{equation}
		is positive, where $E_{ij}$ are the matrix units of  $\mathbb{M}_p$.
	\end{enumerate}

\end{theo}

Theorem~\ref{theo:blockcompletion} makes the connection between PSD block decomposable matrices and completely positive maps. Our partial trace regression formulation is based on learning completely positive maps via low-rank Kraus decomposition, and thus can be applied to the problem of  PSD matrix completion.
The most straightforward application of Theorem~\ref{theo:blockcompletion} to PSD matrix completion is to consider the case where the matrix is block-structured with missing blocks.
This boils down to solving the following optimization problem
 \begin{equation*}
\label{eq:blockmatrixcompletion}
\argmin_{A_k\in\mathbb{M}_{q\times p}} \sum_{i,j=1}^p \big\|Y_{ij} - \Phi(X_{ij})\big\|_F^2 \text{  \ s.t. \ } \Phi (\cdot)=\sum_{k=1}^r A_k \cdot A_k^\top,
\end{equation*}
where $\|\cdot\|_F$ is the Frobenius norm, $Y_{ij}$ are the observed blocks of the matrix and $X_{ij}$ are the corresponding matrix units.
So, a completely positive map $\Phi$ can be learned by our approach from the available blocks and then can be used to predict the missing blocks. This would be a natural approach to take into account local structures in the matrix and then improve the completion performance.
This approach can be also applicable in  situations where no information about the block-decomposability of the  data matrix to be completed is available. In this case, the size of the blocks $q$ and the number of the blocks $p$ can be viewed as hyperparameters of the model, and can be tuned  to fit the data. Note that when $q=1$, our method reduces to the standard trace regression-based matrix completion.


%% file: bound.tex


 \textbf{Generalization. } We now examine the generalization properties of  partial trace regression via low-rank Kraus decomposition. Specifically, using the notion of pseudo-dimension, we provide  an upper bound on the excess-risk for the function class $\mathcal{F}$ of completely postive maps  with low Kraus rank, i.e., 
 \begin{align*}
 \mathcal{F} = \{\Phi:  \mathbb{M}_{p} \to \mathbb{M}_{q} : \Phi  &\text{  is completely  positive and }\\ & \text{its Kraus rank is equal to } r  \}.
 \end{align*}
 Recall that the expected loss of any hypothesis $h\in\mathcal{F}$ is defined by $R(h) =\mathbb{E}_{(X,Y)}\big[\ell\big(Y, h(X)\big)\big]$ and its empirical loss by $\hat{R}(h) =\frac{1}{l}\sum_{i=1}^l\ell\big(Y_i, h(X_i)\big)$.

 The analysis presented here follows the lines of ~\citet{srebro2004learning} 
 in which generalization bounds were derived for low-rank matrix factorization (see also~\citet{rabusseau2016low} where similar results were obtained for low-rank tensor regression). 
 In order to apply known results on pseudo-dimension,  we  consider  the class of real-valued functions  $\tilde{\mathcal{F}}$ with domain $\mathbb{M}_{p}  \times [q] \times [q]$, where $[q]\doteq \{1,\ldots,q\}$, defined by 
 \begin{equation*}
 \tilde{\mathcal{F}} =  \{(X,s,t)\mapsto \big(\Phi(X)\big)_{st} : \Phi(X)  = \sum_{j=1}^{r} A_j X A_j^\top  \}.
 \end{equation*}

\begin{lem}\label{lem:pseudodimension}
	The pseudo-dimension of the real-valued function class $\tilde{\mathcal{F}}$ is upper bounded by $pqr \log(\frac{8epq}{r})$.
\end{lem}

We can now invoke standard generalization error bounds in
terms of the pseudodimension~\citep[Theorem 10.6]{mohri2018foundations} to obtain:
 
 \begin{theo}\label{theo:bound}
 	Let $\ell: \mathbb{M}_q\to\mathbb{R}$ be a loss function
 	satisfying 
 	$$
 	\ell(Y,Y') = \frac{1}{q^2}\sum_{s,t}
 	\ell'(Y_{st},Y'_{st})
 	$$
 	for some loss function $\ell':\mathbb{R}\to\mathbb{R}^+$ bounded by $\gamma$.
 	Then for any $\delta>0$, with probability at least
 	$1-\delta$ over the choice of a sample of size $l$, the following inequality holds for all
 	$h\in\mathcal{F}$:
 	$$
 	R(h) \leq \hat{R}(h) 
 	+ 
 	\gamma \sqrt{\frac{pqr \log(\frac{8epq}{r})\log (\frac{l}{pqr})}{l}}
 	+
 	\gamma \sqrt{\frac{\log\left(\frac{1}{\delta}\right)}{2l}}.
 	$$
 \end{theo}

The proofs of Lemma~\ref{lem:pseudodimension} and Theorem~\ref{theo:bound} are provided in the Supplementary Material.
 Theorem~\ref{theo:bound} shows that the Kraus rank $r$ plays the role of a regularization parameter. Our generalization bound suggests a trade-off between reducing the empirical error which may require a more complex hypothesis set~(large $r$), and controlling the complexity of the hypothesis set which may increase the empirical error~(small~$r$).


%% file: expe.tex


In this section we turn our attention to evaluating our proposed partial trace regression (PTR) method. 
We conduct experiments on two tasks where partial trace regression can help, both in a simulated setting and exploiting real data. 
In all the experiments, the PTR model is implemented in a keras/Tensorflow framework and learned with Adam with default learning rate (0.001) and for 100 epochs. Batch size is typically $16$.
Our code is available at \url{https://github.com/Stef-hub/partial_trace_kraus}.

\subsection{PSD to PSD Matrix Regression}

We will now describe experiments where the learning problem can be easily described as learning a mapping between two PSD matrices; first with simulated data and then applied to learning covariance matrices for Brain-Computer Interfaces.

\subsubsection{Experiments on Simulated Data}

\begin{figure}[tb]
	\centering
	\includegraphics[width=0.43\textwidth]{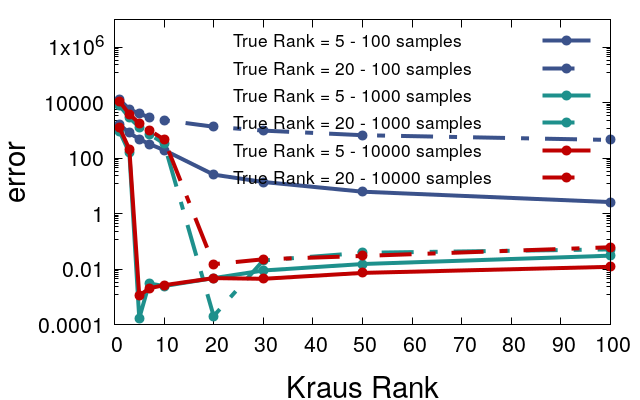} \\ \medskip
	\includegraphics[width=0.43\textwidth]{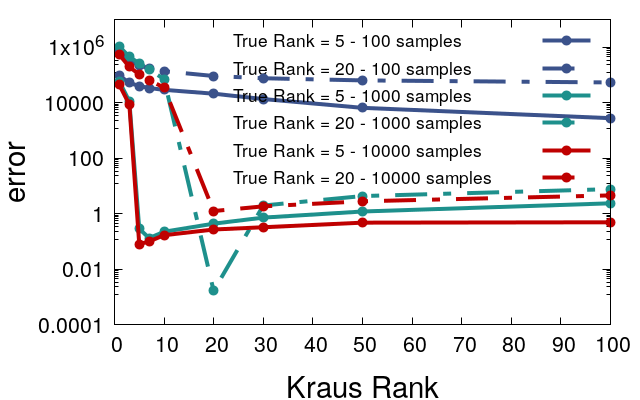}
	\caption{\label{fig:psdsimxp}PSD to PSD predictive performance (mean squared error) of PTR as a function of its Kraus rank, on simulated data from map $20 \times 20 \rightarrow 10 \times 10$ (top) and map $100 \times 100 \rightarrow 40 \times 40$~(bottom).}
\end{figure}

Our first goal is to show the ability of our model to accurately recover mappings conforming to its assumptions. 
We randomly draw a set of matrices $X_i$ and $A_r$, and build the matrices $Y_i$ using Equation~\ref{eq:kraus}, for various Kraus ranks $r$, and $p$, $q$ the size of input and output spaces respectively. We train the model with 100, 1000 and 10000 samples on two simulated datasets with Kraus ranks 5 and 20, and show the results for maps $20 \times 20 \rightarrow 10 \times 10$ and $100 \times 100 \rightarrow 40 \times 40$ in Figure~\ref{fig:psdsimxp}. 
While 100 samples is not enough to get a good estimation, with more data the PTR is able to accurately represent the model with correct rank.

We also aim to show, in this setting, a comparison of our PTR to trace regression and two baselines  that are commonly used in multiple output regression tasks: the multivariate regression where no rank assumptions are made ($y = xB +\epsilon$) and  the reduced-rank multivariate regression ($y = xB +\epsilon$ with a low-rank constraint on $B$). For these two methods, $B$ is a matrix mapping the input $x$ to the output $y$, which are the vectorization of the  matrix-valued  input $X$ and the matrix-valued output $Y$, respectively.
We  also compare PTR to three tensor to tensor regression methods: the higher order partial least squares~(HOPLS)~\citep{zhao2012higher} and the tensor train and the Tucker tensor neural networks (NN)~\citep{novikov2015tensorizing,cao2017tensor}.

All the models are trained using 10000 simulated examples from map $20 \times 20 \rightarrow 10 \times 10$ generated from a model with true Kraus rank 5. The results are shown in Table~\ref{tab:psdsimxp} on 1000 test samples where we display the best performance over ranks from 1 to 100 (when applicable) in terms of mean squared error (MSE). 
For the (reduced-rank) multivariate regression experiments we needed to consider vectorisations of the matrices, thus removing some of the relevant structure of the output data. We note that reduced-rank multivariate regression performs worse than multivariate regression since the rank of multivariate regression is not related to the Kraus rank. 
 In this experiment, PTR performs similar to tensor train NN and significantly better than all the other methods.
 Note that, in contrast to PTR, tensor train NN does not preserve the PSD property.
 
 We ran these experiments also with fewer examples (100 instead of 10000). In this case PTR performs better than tensor train NN ($0.007 \pm 0.013$ for PTR and $0.662 \pm 0.364$ for tensor train NN) and again better than the other baselines.

\begin{table}[t]
	\caption{Comparison of various regression models on PSD to PSD task on simulated data with map $20 \times 20 \rightarrow 10 \times 10$ and true Kraus rank of the model $5$. We report the best performance among various tested model ranks from 1 to 100. \label{tab:psdsimxp}} 
		\begin{center}
\begin{tabular}{lc}
	\hline
	Model & MSE \\\hline
	Partial Trace Regression		& $0.001 \pm 0.0008$ \\ 
	Trace Regression 			& $0.028 \pm 0.0144$ \\
	Multivariate Regression 		& $0.058 \pm 0.0134$ \\
	Reduced-rank Multivariate Regression	&  $0.245 \pm 0.1023$ \\
	HOPLS	&  $1.602 \pm 0.0011$ \\ 
	Tensor Train NN	&  $0.001 \pm 0.0009$ \\
	Tucker NN	&  $0.595 \pm 0.0252$ \\
	\hline 
\end{tabular}
	\end{center}
\end{table}

\subsubsection{Mapping Covariance Matrices for Domain Adaptation}

In some applications like 3D shape recognition or Brain-Computer Interfaces \cite{barachant2011multiclass,tabia2014covariance}, features take the form of covariance matrices and algorithms taking
into account the specific geometry of these objects are needed. For instance,
in BCI, "minimum distance to the mean  (in the Riemannian sense)" classifier has been shown to be
a highly efficient classifier for categorizing motor imagery tasks. 

For these tasks, distribution shifts usually occur in-between sessions of the same subject using the BCI. 
In such situations, one solution is to consider an optimal transport mapping of the covariance matrices from the source to the target domain (the different sessions) \cite{yair2019optimal,courty2016optimal}.  Here, our goal is to learn such a covariance matrix mapping and to perform classification using covariance matrix from one session (the source domain) as 
training data and those of the other session (the target domain) as test data.
For learning the mapping, we will consider only matrices from the training session.
  
We adopt the experimental setting described by~\citet{yair2019optimal} for generating the covariance matrix.  From the  optimal transport-based mapping obtained in a unsupervised way from the training session,  
we have couples of input-output matrices of size $22\times 22$ from which we want to learn our partial trace regressor. While introducing noise into the classification process, the benefit of such regression function is to allow out-of-sample domain adaptation as in \citet{perrot2016mapping}. This means that we do not need to solve an optimal
transport problem every time a new sample (or a batch of new samples) is available.
In practice, we separate all the 
matrices (about $270$) from the training session in a training/test group, and use
the training examples ($230$ samples) for learning the covariance mapping. For evaluating the quality
of the learned mapping, we compare the classification performance of a minimum distance to the mean classifier in three situations:
\begin{itemize}
	\item No adaptation between domains is performed. The training session data is used as is. The method is denoted as NoAdapt.
	\item All the matrices from the training session are mapped using the OT mapping. This is a full adaptation approach, denotes as FullAdapt
	\item Our approach denoted as OoSAdapt (from Out of Sample Adaptation) uses
	the $230$ covariance matrices mapped using OT and the other matrices mapped using our partial trace regression model.  
\end{itemize}
For our approach, we report classification accuracy  for a model of rank 20 and depth $1$ trained using an Adam optimize of learning rate $0.001$ during $500$ iterations. We have also tested several other hypeparameters (rank and depth)
without much variations in the average performance.

Classification accuracy over the test set (the matrice from the second session) is reported in Table \ref{tab:bci}. We first note that for all subjects, domain adaptation
helps in improving performance. Using the estimated mapping instead of the
true one for projecting source covariance matrix into the target domain,
we expect a loss of performance with respect to the FullAdapt method. 
Indeed, we observe that for Subject 1 and 8, we occur small losses. 
For the other subjects, our method allows to improve performance compared
to no domain adaptation while allowing for out-of-sample classification.
Interestingly, for Subject 9, using estimated covariance matrices performs slightly better than using the exact ones.

\begin{table}[t]
	\caption{ Accuracy of a "minimum distance to the mean" classifier on domain adaptation BCI task. We report the results for the same subjects as in \cite{yair2019optimal}. 
	\label{tab:bci}} 
		\begin{center}
\begin{tabular}{lccc}
	\hline
	Subject & NoAdapt & FullAdapt & OoSAdapt \\\hline
	1		& 73.66	  & 73.66 & 72.24 \\
	3 		& 65.93   & 72.89 & 68.86 \\
	7		& 53.42   & 64.62 & 59.20 \\	
	8		& 73.80   & 75.27 & 73.06  \\
	9 		& 73.10   & 75.00 & 76.89 \\\hline  	
\end{tabular}
	\end{center}
\end{table}

\begin{figure}[tb]
	\centering
	\includegraphics[width=0.15\textwidth]{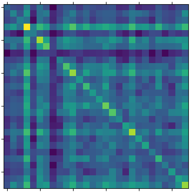} \
	\includegraphics[width=0.15\textwidth]{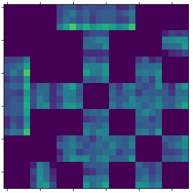} \
	\includegraphics[width=0.15\textwidth]{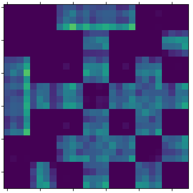} \\ \
	\includegraphics[width=0.15\textwidth]{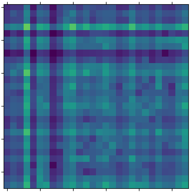} \
	\includegraphics[width=0.15\textwidth]{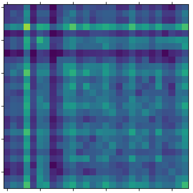} \
	\includegraphics[width=0.15\textwidth]{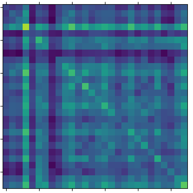} \
	\vspace{-0.2cm}\caption{\label{fig:complblocs}Completion performance on simulated $28 \times 28$ matrix, with $p = 7$ and $q = 4$. Top left, middle and right: original matrix, original matrix with missing values, our result with rank 1 Kraus decomposition. Bottom left, middle and right: Our results with Kraus rank 5, 30 and 100.}
\end{figure}

\begin{table}[t]
	\caption{Comparison of partial trace regression and trace regression and tensor train neural networks on (block) PSD matrix completion on simulated $28 \times 28$ matrix with missing blocks. \label{tab:completionsimxp}} 
	\begin{center}
		\begin{tabular}{lc}
			\hline
			Model & MSE \\\hline
			Partial Trace Regression		& $0.572 \pm 0.019$ \\ 
			Trace Regression 			& $2.996 \pm 1.078$ \\
			Tensor Train NN	&  $3.942 \pm 1.463$ \\
			\hline 
		\end{tabular}
	\end{center}
\end{table}

\subsection{PSD Matrix Completion}

\begin{figure}[tb]
	\centering
	\includegraphics[width=0.15\textwidth]{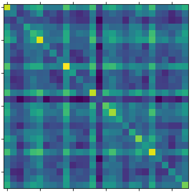} \
	\includegraphics[width=0.15\textwidth]{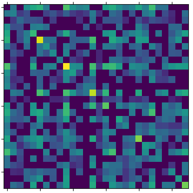} \\ \
	\includegraphics[width=0.15\textwidth]{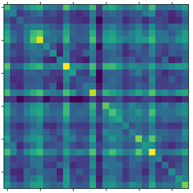} \
	\includegraphics[width=0.15\textwidth]{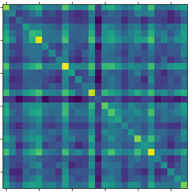} \
	\caption{\label{fig:complvaluesA}Completion performance on simulated $28 \times 28$ matrix, with $p = 7$ and $q = 4$. Top row: target matrix and target matrix with missing data. Bottom row: Our completion results with depth=1 (left) and depth=2 (right).}
\end{figure}

\begin{figure}[tb]
	\centering
	\includegraphics[width=0.15\textwidth]{M_orig_2.png} \
	\includegraphics[width=0.15\textwidth]{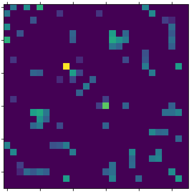} \\ \
	\includegraphics[width=0.15\textwidth]{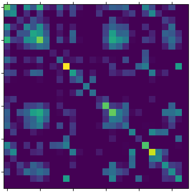} \
	\includegraphics[width=0.15\textwidth]{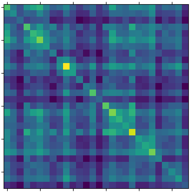} \
	\caption{\label{fig:complvaluesB}Completion performance on simulated $28 \times 28$ matrix, with $p = 7$ and $q = 4$. Top row: target matrix and target matrix with missing data. Bottom row: Our completion results with depth=1 (left) and depth=2 (right).}
\end{figure}

We now describe our experiments in the matrix completion setting, first by illustrative examples with simulated data, then more comprehensively in the setting of multi-view kernel matrix completion.

\subsubsection{Experiments on Simulated Data}

We consider the problem of matrix completion in two settings: filling in fully missing blocks, and filling in individually missing values in a matrix. We perform our experiments on full rank PSD matrices of size $28 \times 28$.  

We show the results on block completion in Figure~\ref{fig:complblocs}, where we have trained our partial trace regression model (without stacking) with Kraus ranks 1, 5, 30 and 100. 
While using Kraus rank 1 is not enough to retrieve missing blocs, rank 5 gives already reasonable results, and ranks 30 and 100 are able to infer diagonal values that were totally missing from the training blocks. 
Completion performance in terms of mean squared error are reported in Table~\ref{tab:completionsimxp}, showing that our PTR method performs favorably against trace regression and tensor train neural networks.

Figures \ref{fig:complvaluesA} and \ref{fig:complvaluesB} illustrate the more traditional completion task where $35\%$ and $85\%$ (respectively) of values in the matrix are missing independently of the block structure. 
We fix $p$ and $q$ to 7 and 4, respectively, and investigate the effect that stacking the models has on the completion performance. Note that $p$ and $q$ may be chosen via cross-validation.
With only a little missing data (Figure~\ref{fig:complvaluesA}) there is very little difference between the results obtained with one-layer and two-layer models.
However we observe that for the more difficult case (Figure~\ref{fig:complvaluesB}), stacking partial trace regression models significantly improves  the reconstruction performance.

\begin{figure}[tb]
	\centering
	~\hfill
	\includegraphics[width=0.34\textwidth]{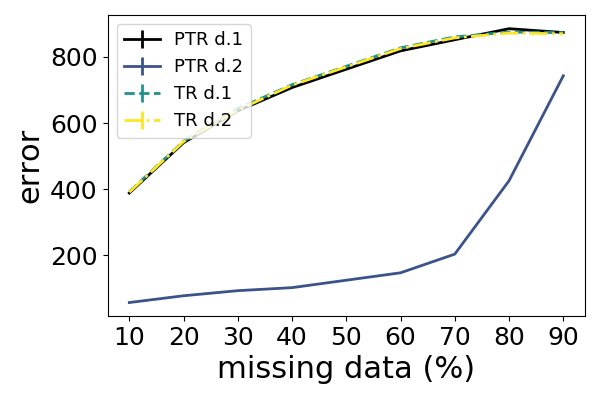} ~\hfill
	\caption{\label{fig:svm_completion}Sum of the matrix completion errors for trace regression~(TR) and our partial trace regression~(PTR), over the views of the multiple features digits dataset as a function of missing samples with model depths 1 and 2. }
\end{figure}

\begin{figure*}[t]
	\centering
	\includegraphics[width=0.31\textwidth]{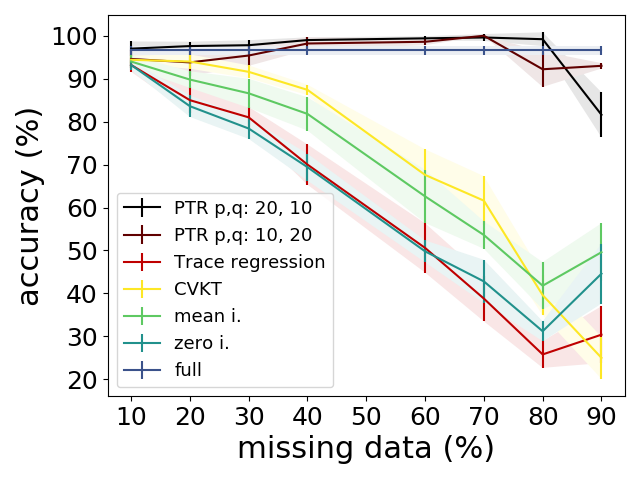} 
	\includegraphics[width=0.31\textwidth]{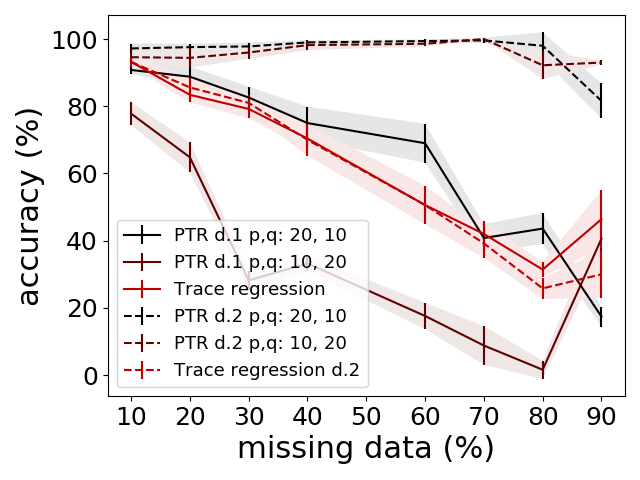}
	\includegraphics[width=0.31\textwidth]{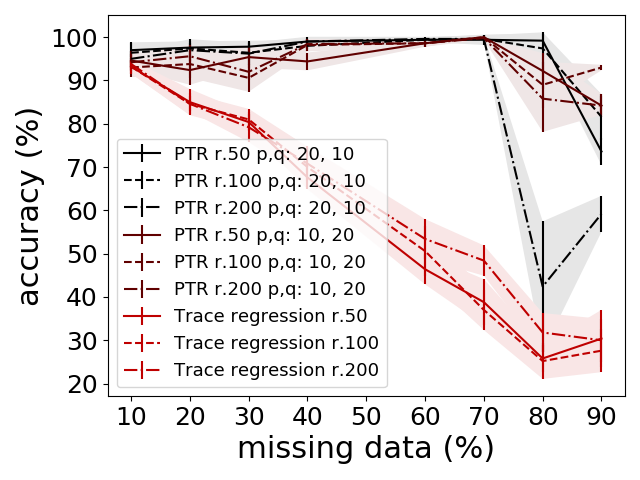}
	\caption{\label{fig:svm}SVM accuracy results on the digits dataset as functions of the amount of missing data samples. Left: the classification accuracies with kernel matrices completed with the compared methods and with "full" kernel matrices for reference; Middle: The results of our method separately w.r.t. the depth of the model;  Right: The results of our method separately w.r.t the assumed Kraus rank. }
\end{figure*}

\subsubsection{Similarity Matrix Completion}
For our last set of experiments, we evaluate our partial trace regression model in the task of matrix completion on a multi-view dataset, where each view contains missing samples. This scenario occurs in many fields where the data comes from various sensors of different types producing heterogeneous view data~\cite{rivero2017mutual}. Following~\citet{huusari2019kernel}, we consider the multiple features digits dataset, available online.\footnote{https://archive.ics.uci.edu/ml/datasets/Multiple+Features} The task is to classify digits (0 to 9) from six descriptions extracted from the images, such as Fourier coefficients or Karhunen-Loéve coefficients. We selected 20 samples from all the 10 classes, and computed RBF kernels for views with data samples in $\mathbb{R}^d$, and $Chi^2$ kernels for views with data samples in $\mathbb{Z}^d$, resulting in six $200 \times 200$ kernel matrices. We then randomly introduced missing samples within views, leading to missing values (rows/columns) into kernel matrices. We vary the level of total missing samples in the whole dataset from 10\% to 90\%, by taking care that all the samples are observed at least in one view, and that all views have observed samples. 

We first measure the matrix completion performance on reconstruction quality by computing $\sum_v \|K_v - \hat{K}_v \|_F$ with $K_v$ the original kernel matrix for view $v$ and $\hat{K}_v$ the completed one. We then analyse the success of our method in the classification task by learning an SVM classifier from a simple average combination of input kernels. Here we compare our partial trace regression method to two very simple baselines of matrix completion, namely zero and mean inputation, as well as the more relevant CVKT method presented in~\citet{huusari2019kernel}. We perform the matrix completion with our algorithm with three block-structure configurations; $p, q=20,10$, $p, q=10,20$, and finally with $p, q=200,1$, which corresponds to trace regression. We consider both depths 1 and 2 for our partial trace regression model, as well as Kraus ranks 50, 100 and 200.

Figure~\ref{fig:svm_completion} shows the sum of matrix completion errors $\sum_v \|K_v - \hat{K}_v\|_F$ for our method and the trace regression method in various configurations. For depth 2, the partial trace regression clearly outperforms the more simple trace regression ($p, q=200,1$). 
With model depth 1 all the methods perform similarly.
It might be that the real data considered in this experiment is more complex and non-linear than our model assumes, thus stacking our models is useful for the performance gain. However the traditional trace regression does not seem to be able to capture the important aspects of this data even in the stacked setting.

Figure~\ref{fig:svm} shows in the left panel the SVM classification accuracies obtained by using the kernel matrices from various completion methods, and detailed results focusing on our method in the middle and right side panels. For these experiments, we selected to use in SVMs the kernel matrices giving the lowest completion errors.   
We observe that our model provides excellent kernel completion since the classification accuracy is close to the performance of the setting with no missing data. The stacked model seems to be able to accurately capture the data distribution, giving rise to very good classification performance. 
The middle panel confirms the observations from Figure~\ref{fig:svm_completion}: the depth of our model plays a crucial part on its performance, with depth 2 outperforming the depth 1 in almost every case, except the choice of $p,q=200,1$ corresponding to trace regression.  
The Kraus rank does not have a strong effect on classification performance (right panel). Indeed, this justifies the usage of our method in a low-rank setting.

%% file: conclusion.tex

In this paper, we introduced {\em partial trace regression} model, a family of linear predictors from matrix-valued inputs to matrix-valued outputs that generalizes previously proposed models such as  trace regression and  linear regression. We proposed a novel framework for estimating the partial trace regression model from data by learning  low-rank Kraus decompositions of completely positive maps, and derived an upper bound on the generalization error. Our empirical study with synthetic and real-world datasets  shows the promising performance of our proposed approach on the tasks of matrix-to-marix regression and positive semidefinite matrix completion.

%% file: appendix.tex

In this supplementary material, we prove Lemma~3 and Theorem~4 in Section~2.3 of the main paper.
Let us first recall the definition of  pseudo-dimension.
\begin{de}\label{de:Shattering} (Shattering~\citealp[Def.~10.1]{mohri2018foundations}) \\[0.1cm] 
	Let $G$ be a family of functions from $X$ to $\mathbb{R}$. A set $\{x_1, . . . , x_m\} \subset X$ is said to be shattered by $G$ if there exist $t_1, . . . , t_m \in \mathbb{R}$ such that,
	\begin{equation*}
	f(x)=
	\left|\left\lbrace\left[
	\begin{array}{c}
	 sign(g(x_1)-t_1)\\
	 \vdots \\
	 sign(g(x_m)-t_m)
	\end{array}\right] :   g \in G  \right\rbrace\right| = 2^m.
	\end{equation*}
\end{de}
\begin{de}\label{de:pseudodimension} (pseudo-dimension~\citealp[Def.~10.2]{mohri2018foundations}) \\[0.1cm] 
	Let $G$ be a family of functions from $X$ to $\mathbb{R}$. Then, the pseudo-dimension
of $G$, denoted by $Pdim(G)$, is the size of the largest set shattered by $G$.
\end{de}
In the following we consider that the expected loss of any hypothesis $h\in\mathcal{F}$ is defined by $R(h) =\mathbb{E}_{(X,Y)}\big[\ell\big(Y, h(X)\big)\big]$ and its empirical loss by $\hat{R}(h) =\frac{1}{l}\sum_{i=1}^l\ell\big(Y, h(X)\big)$.
To prove Lemma~3 and Theorem~4,  we need the following two results. 

\begin{theo}\label{theo:sign}{\citep[Theorem 35]{srebro2004learning}}\\[0.1cm]
	The number of sign configurations of $m$ polynomials, each of degree at most $d$, over $n$ variables is at most
	$\left(\frac{4edm}{n}\right)^n$
	for all $m>n>2$.
\end{theo}

\begin{theo}\label{theo:generalization}{\citep[Theorem 10.6]{mohri2018foundations}}\\[0.1cm]
	Let $H$ be a family of real-valued functions and let $G = \{x \mapsto L(h(x), f (x)) : h \in H\}$
	be the family of loss functions associated to $H$. Assume that the pseudo-dimension
	of $G$ is bounded by $d$ and that
	the loss function $L$ is bounded by $M$. Then, for any $\delta>0$, with probability at least
	$\delta$ over the choice of a sample of size $m$, the following inequality holds for all
	$h \in H$:
	
	$$
	R(h) \leq \hat{R}(h) 
	+ 
	M \sqrt{\frac{2d
			\log\left(\frac{em}{d}\right)}{m}}
	+
	M \sqrt{\frac{\log\left(\frac{1}{\delta}\right)}{2m}}.
	$$
\end{theo}

\section{Proof of Lemma~3}
We now prove Lemma~3 in Section~2.3 of the main paper.

\setcounter{theo}{2}
\begin{lem}\label{lem:pseudodimension}
	The pseudo-dimension of the real-valued function class $\tilde{\mathcal{F}}$ with domain $\mathbb{M}_{p}  \times [q] \times [q]$ defined by 
	\begin{equation*}
	\tilde{\mathcal{F}} =  \{(X,s,t)\mapsto \big(\Phi(X)\big)_{st} : \Phi(X)  = \sum_{j=1}^{r} A_j X A_j^\top  \}
	\end{equation*}
	is upper bounded by $pqr \log(\frac{8epq}{r})$.
\end{lem}

\textbf{Proof:}
It is well known that the pseudo-dimension of a vector space of real-valued functions 
is equal to its dimension~\citep[Theorem~10.5]{mohri2018foundations}. Since $\tilde{\mathcal{F}}$ is 
a subspace of the $p^2q^2$-dimensional vector space 
$$ \left\{(X,s,t)\mapsto \big(\Phi(X)\big)_{st}\ :\ 
\Phi \in \mathcal{L}(\mathbb{M}_{p};\mathbb{M}_{q})  \right\}$$
of real-valued functions  with domain $\mathbb{M}_{p}  \times [q] \times [q]$
the pseudo-dimension of $\tilde{\mathcal{F}}$ is bounded by $p^2q^2$. 

Now, let $m\leq p^2q^2$ and let $\{(X_k,s_k,t_k)\}_{k=1}^m$ be 
a set of points that are pseudo-shattered by $\tilde{\mathcal{F}}$ with thresholds $t_1,\cdots,t_m\in\mathbb{R}$.
Then for each binary labeling $(u_1,\cdots,u_m)\in\{-,+\}^m$, there exists $\tilde{\Phi}\in\tilde{\mathcal{F}}$ such
that $sign(\tilde{\Phi}(X_k,s_k,t_k) - v_k) = u_k$. Any function $\tilde{\Phi}\in\tilde{\mathcal{F}}$
can be written as  
\begin{equation}
\label{eq:tildephi}
\tilde{\Phi}(X,s,t) = \big(\sum_{j=1}^{r} A_j X A_j^\top\big)_{st}, 
\end{equation}
where $A_j \in \mathbb{M}_{q\times p}, \forall j\in[r]$.
If we consider the $pqr$ entries of  $A_j$, $j =1,\ldots,r$, as variables,  the set
$\{\tilde{\Phi}(X_k,s_k,t_k) - v_k\}_{k=1}^m$ can be seen (using Eq.~\ref{eq:tildephi}) as a set of $m$ polynomials of degree $2$ over these variables.
Applying  Theorem~\ref{theo:sign} above, we obtain that the number of  sign configurations, which is equal to $2^m$, is bounded by $\left(\frac{8em}{pqr}\right)^{pqr}$.  The result follows since $m\leq p^2q^2$.
\hfill $\blacksquare$

\section{Proof of Theorem~4}
In this section, we prove Theorem~4 in Section~2.3 of the main paper.

 \begin{theo}\label{theo:bound}
	Let $\ell: \mathbb{M}_q\to\mathbb{R}$ be a loss function
	satisfying 
	$$
	\ell(Y,Y') = \frac{1}{q^2}\sum_{s,t}
	\ell'(Y_{st},Y'_{st})
	$$
	for some loss function $\ell':\mathbb{R}\to\mathbb{R}^+$ bounded by $\gamma$.
	Then for any $\delta>0$, with probability at least
	$1-\delta$ over the choice of a sample of size $l$, the following inequality holds for all
	$h\in\mathcal{F}$:
	$$
	R(h) \leq \hat{R}(h) 
	+ 
	\gamma \sqrt{\frac{pqr \log(\frac{8epq}{r})\log (\frac{l}{pqr})}{l}}
	+
	\gamma \sqrt{\frac{\log\left(\frac{1}{\delta}\right)}{2l}}.
	$$
\end{theo}

\textbf{Proof:}
For any $h:\mathbb{M}_{p} \to \mathbb{M}_{q} $ we define
$\tilde{h}:\mathbb{M}_{p}  \times [q] \times [q] \to \mathbb{R}$ by
$\tilde{h}(X,s,t) = \big(h(X)\big)_{st}$.
Let $\mathcal{D}$ denote the distribution of the input-output data. We have
\begin{align*}
R(h) 
&= 
\mathbb{E}_{(X,Y)\sim \mathcal{D}} [\ell(Y,h(X))]\\
&= 
\frac{1}{q^2}\sum_{s,t}\mathbb{E}_{(X,Y)\sim \mathcal{D}} [\ell'(Y_{st},h(X)_{st}]\\
&=
\mathbb{E}_{(X,Y) \sim \mathcal{D} \atop s,t \sim \mathcal{U}(q)} 
[\ell'(Y_{st},\tilde{h}(X,s,t))],
\end{align*}
where $\mathcal{U}(q)$ denotes the discrete uniform distribution on $[q]$. It follows that $R(h) = R(\tilde{h})$. By the same way, we can
show that $\hat{R}(h) = \hat{R}(\tilde{h})$. The generalization bound is then obtained  using Theorem~\ref{theo:generalization} above.
\hfill $\blacksquare$
